%% file: main_paper.tex
\documentclass[conference]{IEEEtran}
\IEEEoverridecommandlockouts
\usepackage{cite}
\usepackage{amsmath,amssymb,amsfonts}
\usepackage{algorithmic}
\usepackage{graphicx}
\usepackage{textcomp}
\usepackage{xcolor, color}
\usepackage{multirow}
\usepackage{subcaption}
\usepackage{booktabs}
\usepackage{makecell}
\def\BibTeX{{\rm B\kern-.05em{\sc i\kern-.025em b}\kern-.08em
    T\kern-.1667em\lower.7ex\hbox{E}\kern-.125emX}}

\begin{document}

\title{Spatial-Temporal Networks for \\Antibiogram Pattern Prediction}

\makeatletter
\newcommand{\linebreakand}{%
  \end{@IEEEauthorhalign}
  \hfill\mbox{}\par
  \mbox{}\hfill\begin{@IEEEauthorhalign}
}
\makeatother

\author{\IEEEauthorblockN{Xingbo Fu}
\IEEEauthorblockA{\textit{University of Virginia}\\
Charlottesville, USA \\
xf3av@virginia.edu}
\and
\IEEEauthorblockN{Chen Chen}
\IEEEauthorblockA{\textit{University of Virginia}\\
Charlottesville, USA \\
zrh6du@virginia.edu}
\and
\IEEEauthorblockN{Yushun Dong}
\IEEEauthorblockA{\textit{University of Virginia}\\
Charlottesville, USA \\
yd6eb@virginia.edu}
\and
\IEEEauthorblockN{Anil Vullikanti}
\IEEEauthorblockA{\textit{University of Virginia}\\
Charlottesville, USA \\
vsakumar@virginia.edu}
\linebreakand
\IEEEauthorblockN{Eili Klein}
\IEEEauthorblockA{\textit{Johns Hopkins University}\\
Baltimore, USA \\
eklein8@jhmi.edu}
\and
\IEEEauthorblockN{Gregory Madden}
\IEEEauthorblockA{\textit{University of Virginia}\\
Charlottesville, USA \\
grm7q@virginia.edu}
\and
\IEEEauthorblockN{Jundong Li}
\IEEEauthorblockA{\textit{University of Virginia}\\
Charlottesville, USA \\
jundong@virginia.edu}
}

\maketitle

\begin{abstract}
An antibiogram is a periodic summary of antibiotic resistance results of organisms from infected patients to selected antimicrobial drugs. Antibiograms help clinicians to understand regional resistance rates and select appropriate antibiotics in prescriptions. 
In practice, significant combinations of antibiotic resistance may appear in different antibiograms, forming antibiogram patterns. Such patterns may imply the prevalence of some infectious diseases in certain regions. Thus it is of crucial importance to monitor antibiotic resistance trends and track the spread of multi-drug resistant organisms. In this paper, we propose a novel problem of antibiogram pattern prediction that aims to predict which patterns will appear in the future. Despite its importance, tackling this problem encounters a series of challenges and has not yet been explored in the literature. 
First of all, antibiogram patterns are not i.i.d as they may have strong relations with each other due to genomic similarities of the underlying organisms. 
Second, antibiogram patterns are often temporally dependent on the ones that are previously detected. 
Furthermore, the spread of antibiotic resistance can be significantly influenced by nearby or similar regions. 
To address the above challenges, we propose a novel Spatial-Temporal Antibiogram Pattern Prediction framework, STAPP, that can effectively leverage the pattern correlations and exploit the temporal and spatial information.
We conduct extensive experiments on a real-world dataset with antibiogram reports of patients from 1999 to 2012 for 203 cities in the United States. The experimental results show the superiority of STAPP against several competitive baselines.

\end{abstract}


\section{Introduction}
The ever-increasing spread of antibiotic resistance has become a worrisome public health problem around the world \cite{bonhoeffer1997evaluating}. It not only compromises the effectiveness of antibiotics and increases the cost of treatment, but can also transmit between patients and regions. In response to the spread of antibiotic resistance, a number of measures have been proposed, and antibiograms are one of the most prevalent tools adopted by many clinicians for detecting and describing antibiotic resistance. At the patient level, an antibiogram report is a periodic profile of antibiotic resistance testing results from a pathogen cultured from patient samples (e.g., pus from a wound and blood culture); a battery of resistance tests are performed by the microbiology laboratory for drugs representing key antimicrobial classes for that organism and reported to the ordering clinician~\cite{joshi2010antibiogram}. An example of an antibiogram report from a patient is shown in Table~\ref{table:example}. The collection of antibiograms of the patients within a region provides a critical window to assessing regional epidemiology of resistance and informing empirical antimicrobial treatment \cite{truong2021antibiogram}. 

\begin{figure} [t]
\centering
\includegraphics[width=\linewidth]{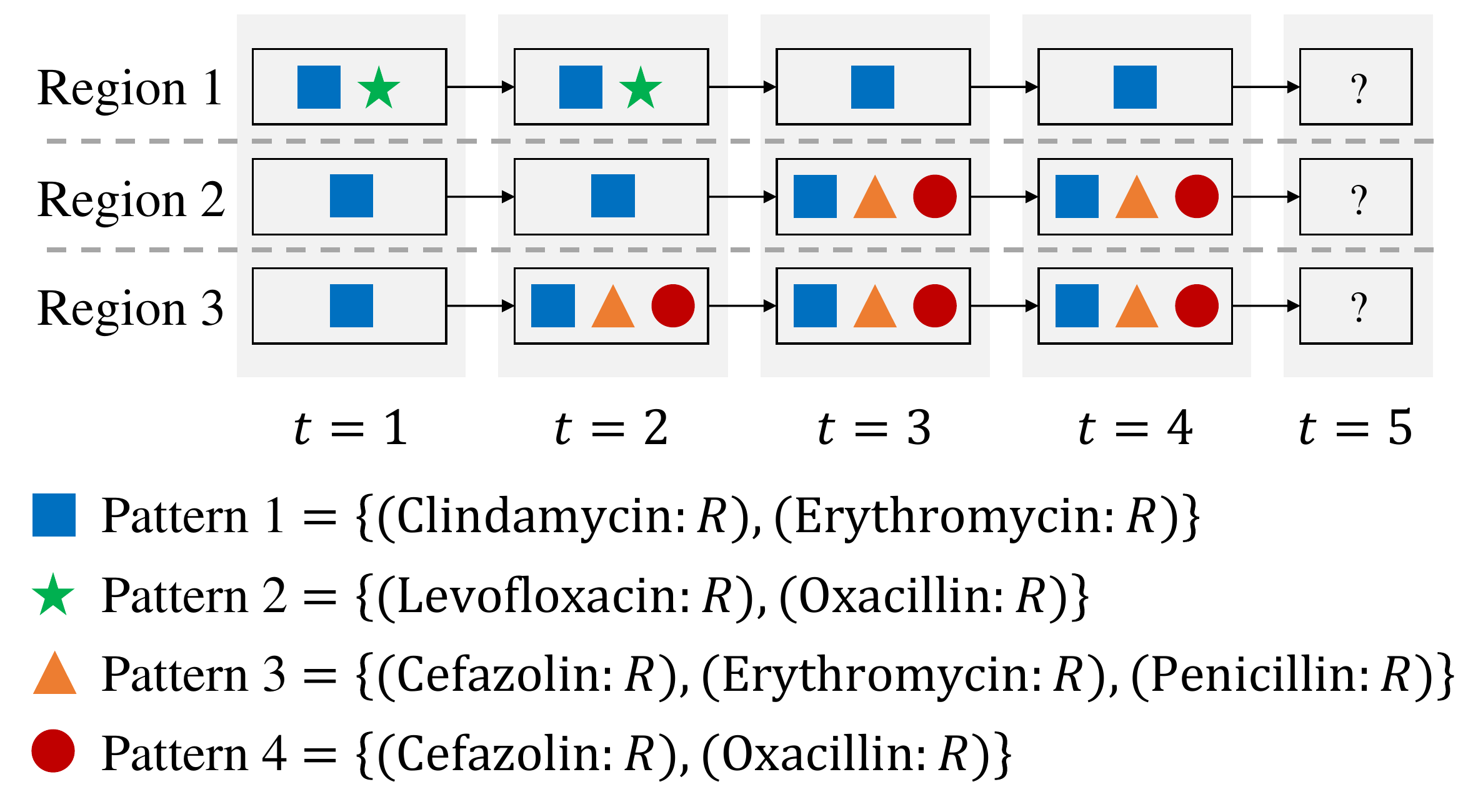}
\caption{An example of antibiogram pattern prediction. This example includes antibiogram patterns appearing in four timesteps from $t=1$ to $t=4$ for three regions. The blue square, the green star, the orange triangle, and the red circle represent four distinct antibiogram patterns. The goal of the three regions is to predict which antibiogram patterns will appear in timestep $t=5$.}
\label{fig:example}
\end{figure}

\input{table_example.tex}

Recently, a number of studies have investigated antibiotic resistance trends over time based on antibiograms. For instance, some studies have analyzed spatial temporal trends in antibiotic resistance \cite{klein:aje13}, and others have attempted to predict antibiotic resistance using machine learning approaches \cite{feretzakis2021machine,jimenez2020feature,lewin2021predicting}.
However, these studies either only consider high level trends or only focus on analyzing resistance trends of one antibiotic individually and ignore the dependencies among antibiotic resistance. 
It is worth noting that in recent years, combinations of antibiotic resistance have emerged in antibiogram reports because of the spread of multi-drug resistant pathogens \cite{siegel2007mdro,lee2019mdro}. For example, methicillin-resistant \emph{Staphylococcus aureus} (MRSA), while defined by its resistance to anti-\emph{staphylococcal} penicillins including methicillin, commonly carries intrinsic resistance to many other antibiotics including cephalosporins, streptomycin, tetracycline, and erythromycin \cite{lee2018mrsa,boucher2008mrsa}. Further, specific MRSA strains may be characterized by unique resistance patterns \cite{10.1038/s41579-018-0147-4}. If such a combination of antibiotic resistance is observed significantly from antibiogram reports for a region, we may say it forms an \emph{antibiogram pattern} for that region. Taking the antibiogram report in Table~\ref{table:example} as an example, we may observe a combination of antibiotic resistance \{(Ciprofloxacin: \emph{R}), (Erythromycin: \emph{R}), (Tetracycline: \emph{R})\} from the antibiogram report. If this combination is significantly detected from different antibiogram reports for a region, it will be regarded as an antibiogram pattern for that region. With extracted antibiogram patterns, we can track the spread of multi-drug resistant organisms like MRSA and monitor antibiotic resistance trends for different regions. Therefore, it is of great importance to perform an elaborate analysis of antibiogram patterns.

In this work, we propose a novel problem of antibiogram pattern prediction. The goal of antibiogram pattern prediction is to predict which antibiogram patterns will appear in the future based on the antibiogram patterns detected previously for different regions. An example of this problem is shown in Fig.~\ref{fig:example}. Currently, this problem is unexplored and faces a series of challenges. 
The first challenge is to model relations between antibiogram patterns effectively. This challenge is derived from the fact that some antibiogram patterns (e.g., Pattern 3 and Pattern 4 in Fig.~\ref{fig:example}) may be more likely to co-occur than other antibiogram patterns in practice. Therefore, we need to involve such relations between antibiogram patterns in the prediction framework. 
Second, capturing temporal dependencies of antibiogram patterns is also important. For a specific region, the antibiogram patterns in the future are likely dependent on those that appeared previously. For example, we can observe that Pattern 1, Pattern 3, and Pattern 4 appear from $t=2$ to $t=4$ for Region 3 in Fig.~\ref{fig:example} and these three patterns may also appear in $t=5$ for Region 3. Therefore, involving temporal dependencies will significantly benefit antibiogram pattern prediction. 
Last but not least, due to the spread of resistant pathogens among regions, inter-region correlations should also be taken into account. For instance, Region 2 and Region 3 in Fig.~\ref{fig:example} have mostly similar antibiogram patterns (e.g., in $t=1, 3, 4$) while Region 1 has quite different antibiogram patterns. A prediction framework is supposed to properly leverage such inter-region correlations for making predictions. 

To this end, we propose a novel Spatial-Temporal Antibiogram Pattern Prediction (STAPP) framework in this study. To the best of our knowledge, STAPP is the first framework for the problem of antibiogram pattern prediction. In STAPP, we first construct antibiogram pattern graphs in different timesteps for each region. We model relations between antibiogram patterns based on their similarities from the historical data. Then STAPP employs an antibiogram pattern graph convolution module to aggregate information via relations between antibiogram patterns. In addition, a temporal attention module is deployed to capture temporal dependencies within a region. Considering the spread of antibiotic resistance among regions, STAPP involves a spatial graph convolution module to model inter-region spatial correlations. To validate the effectiveness of the proposed framework, we conduct extensive experiments on a real-world dataset with patient antibiogram reports including \emph{Staphylococcus aureus} susceptibilities to 22 distinct drugs from 203 cities in the United States from 1999 to 2012. The experimental results demonstrate the superiority of the proposed framework against other baseline algorithms.

Overall, the main contributions of this work are summarized as follows:
\begin{itemize}
    \item \textbf{Problem Formulation.} We study a novel problem of antibiogram pattern prediction and present a formal definition of this problem.
    
    \item \textbf{Algorithmic Design.} We propose a novel framework STAPP for the problem. STAPP first models relations between antibiogram patterns via an antibiogram pattern graph convolution module. Then STAPP employs a temporal attention module to capture intra-region temporal dependencies. In addition, a spatial graph convolution module is adopted to extract inter-region spatial correlations for antibiogram pattern prediction.
    
    \item \textbf{Experimental Evaluations.} We validate the effectiveness of the proposed framework through extensive experiments and provide in-depth analysis of the experiment results.
\end{itemize}

\section{Problem Statement} 
\subsection{Preliminary} 
Before we present the problem definition of antibiogram pattern prediction, we first introduce basic concepts in antibiogram pattern mining and Graph Neural Networks (GNNs). Table~\ref{table:notations} shows the notations and their definitions (or descriptions) adopted in this study.

\subsubsection{Antibiogram Pattern Mining.} 
The aim of antibiogram pattern mining is to find the most significant combinations of antibiotic resistance in a region. 
Suppose that we have a set of $M$ distinct antibiotics $\mathcal{B}=\{b_1, b_2, \cdots, b_M\}$ and a cohort of patients $\mathcal{C}_r=\{c_1, c_2, \cdots\}$ in a region $r$. Each patient $c \in \mathcal{C}_r$ is associated with an antibiogram report which displays antimicrobial susceptibility testing results to the antibiotics in $\mathcal{B}$. An example of an antibiogram report from a patient is shown in Table~\ref{table:example}. The result of an antibiotic is one of \emph{NULL} (unknown), \emph{R} (resistant), and \emph{S} (sensitive). Since \emph{NULL} indicates the unknown resistance state of an antibiotic, we only consider the resistance states \emph{R} and \emph{S} in this study. An antibiogram pattern $\mathcal{P}$ is a significant combination of antibiotic resistance in a period for a region. The antibiogram pattern is in the form of $\{B_1, B_2, \cdots\}$ where each element $B$ represents a single antibiotic $b \in \mathcal{B}$ with its resistance state \emph{R} or \emph{S}.

\input{table_notations.tex}

Directly applying frequent pattern mining methods is a simple way to extract frequent combinations from antibiogram reports \cite{fournier2017fim}.
However, it is prone to involve redundant antibiotics whose states are consistently \emph{R} or \emph{S} for most patients. In addition, these methods also miss significant patterns which do not appear very frequently.
In this study, we choose an alternative approach that extracts dependency rules first and converts them into antibiogram patterns. 
A dependency rule is an implication in the form of $\mathcal{S}_a \rightarrow B$, where the antecedent $\mathcal{S}_a=\{B_1, B_2, \cdots\}$ is a set of antibiotics $\mathcal{B}_a=\{b_1, b_2, \cdots\}$ with their resistance states. 
The consequence $B$ is a single antibiotic $b \in \mathcal{B} \setminus \mathcal{B}_a$ with its resistance state \emph{R} or \emph{S}. 
Typically, the dependency rule $\mathcal{S}_a \rightarrow B$ indicates combinations of the antibiotics in $\mathcal{B}_a$  with their resistance states can cause the resistance state $R$ or $S$ of $b$ given the antibiogram reports of the patients $\mathcal{C}_r$ in region $r$.
In this work, we resort to the Kingfisher algorithm \cite{hamalainen2012kingfisher}, an efficient algorithm which searches for the best non-redundant dependencies, to extract dependency rules.
Compared with traditional frequency-based methods (e.g., Apriori \cite{agrawal1994apriori}, Eclat \cite{zaki2000eclat}, and FP-growth\cite{han2000fpgrowth}), Kingfisher does not have the restrictions like minimum frequency thresholds \cite{mahmood2014infrequent}. Readers may refer to paper \cite{hamalainen2012kingfisher} for more details about the Kingfisher algorithm.

Given antibiogram reports of patients in a region, the Kingfisher algorithm extracts significant dependency rules of antibiotic resistance and each of them is associated with its $p$-value. Typically, the smaller the $p$-value is, the more important the dependency rule is. For each dependency rule $\mathcal{S}_a \rightarrow B$ extracted from Kingfisher, we use the union $\mathcal{S}_a \cup \{B\}$ as an antibiogram pattern $\mathcal{P}$.

\subsubsection{Graph Neural Networks.}
We denote an attributed graph as $\mathcal{G}=(\mathcal{V}, \mathcal{E}, \mathbf{X})$, where $\mathcal{V}=\{v_1, v_2, \cdots, v_N\}$ is the set of $N$ nodes ($N=|\mathcal{V}|$), $\mathcal{E}$ is the edge set, and $\mathbf{X}\in \mathbb{R}^{N\times d}$ is the node attribute matrix. Here $d$ is the number of node attributes. The edges describe the relations between nodes and can also be represented by an adjacency matrix $\mathbf{A}\in \mathbb{R}^{N\times N}$. Therefore, we can also use $\mathcal{G}=(\mathbf{A}, \mathbf{X})$ to denote a graph.
A GNN model $f$ parameterized by $\theta^f$ learns the node embeddings $\mathbf{Z}\in \mathbb{R}^{N\times d_1}$ based on the node attribute matrix $\mathbf{X}$ and the adjacency matrix $\mathbf{A}$ through
\begin{equation}
   \mathbf{Z} = f(\mathbf{X}, \mathbf{A};\theta^f),
\end{equation}
where $d_1$ is the dimension of the node embedding. 

\begin{figure*} [t]
\centering
\includegraphics[width=\linewidth]{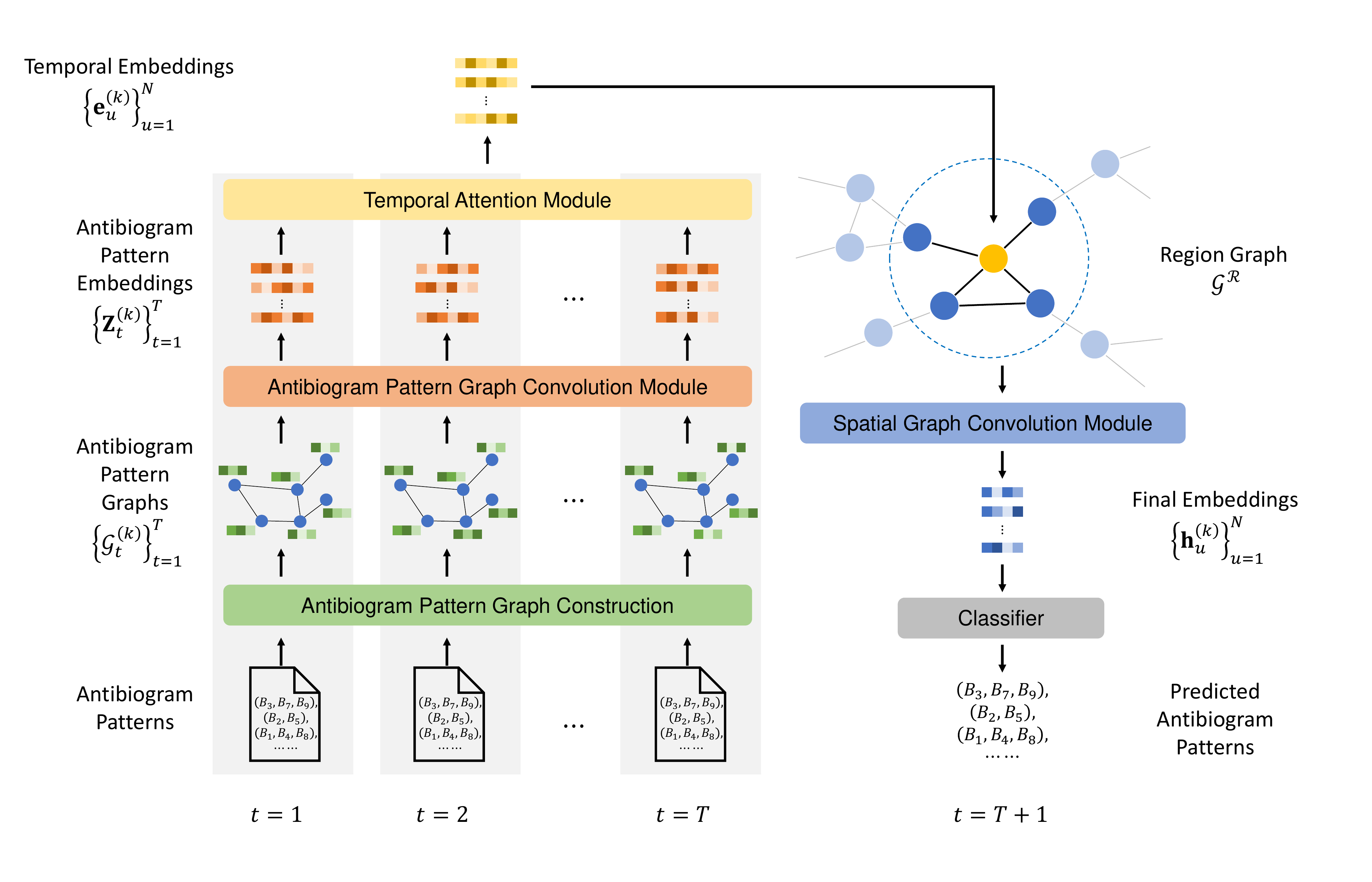}
\caption{An overview of STAPP. STAPP first constructs antibiogram pattern graphs $\{\mathcal{G}^{(k)}_t\}_{t=1}^T$ from $t=1$ to $t=T$ for each region $r^{(k)}$. It then employs an antibiogram pattern graph convolution module, a temporal attention module, and a spatial graph convolution module to predict antibiogram pattern presences in $t=T+1$ using the information from $\{\mathcal{G}^{(k)}_t\}_{t=1}^T$ as well as $r^{(k)}$'s neighboring regions.}
\label{fig:overview}
\end{figure*}

\subsection{Problem Definition}
Based on the aforementioned concepts, we propose to study a novel problem of \textit{Antibiogram Pattern Prediction}, and we formally define the problem as follows.\\

\textsc{Problem 1}. \textbf{(Antibiogram Pattern Prediction.)} \textit{Given a set of $K$ regions $\mathcal{R}=\{r^{(k)}\}_{k=1}^K$, each region $r^{(k)}$ has its antibiogram pattern sets $\{\mathcal{PS}^{(k)}_t\}_{t=1}^T$ for $T$ timesteps. Each antibiogram pattern set $\mathcal{PS}^{(k)}_t=\{\mathcal{P}^{(k)}_{t,1}, \mathcal{P}^{(k)}_{t,2}, \mathcal{P}^{(k)}_{t,3}, \cdots \}$ includes the antibiogram patterns detected in timestep $t$.
The goal is to predict which antibiogram patterns will appear in the next timestep $T+1$ for each region $r^{(k)}$.}\\

\noindent \textbf{Example.} Considering the example in Fig.~\ref{fig:example} with $3$ regions $\mathcal{R}=\{r^{(1)}, r^{(2)}, r^{(3)}\}$, each region $r^{(k)}$ has its antibiogram pattern sets $\{\mathcal{PS}^{(k)}_t\}_{t=1}^4$ from $t=1$ to $t=4$. For instance, $\mathcal{PS}^{(1)}_2 = \{\text{Pattern 1}, \text{Pattern 2}\}$ includes the two antibiogram patterns in $t=2$ for $r^{(1)}$ and $\mathcal{PS}^{(3)}_4 = \{\text{Pattern 1}, \text{Pattern 3}, \text{Pattern 4}\}$ includes the three antibiogram patterns in $t=4$ for $r^{(3)}$. Our goal is to predict which antibiogram patterns will appear in $t=5$ for the three regions.


\section{Methodology} \label{method}
In this section, we elaborate the details of our proposed STAPP, a novel framework tailored for the problem of antibiogram pattern prediction. Fig.~\ref{fig:overview} shows the overview of STAPP. The goal of STAPP is to predict which antibiogram patterns will appear in timestep $t=T+1$ for region $r^{(k)}$ given antibiogram patterns detected from $t=1$ to $t=T$ in region $r^{(k)}$ as well as its neighboring regions. To achieve this goal, STAPP first constructs antibiogram pattern graphs $\{\mathcal{G}^{(k)}_t\}_{t=1}^T$ from $t=1$ to $t=T$ for region $r^{(k)}$. An antibiogram pattern graph convolution module is then employed to embed antibiogram patterns in each $\mathcal{G}^{(k)}_t$. To capture temporal dependencies for antibiogram patterns in different timesteps, we involve a temporal attention module using the attention mechanism in STAPP. Considering the spread of antibiotic resistance among regions, STAPP includes an inter-region spatial graph convolution module to extract spatial correlations among regions. Finally, a classifier module outputs binary predictions indicating which antibiogram patterns will appear in $t=T+1$.

\subsection{Antibiogram Pattern Graph Construction}
Instead of merely leveraging information of a target antibiogram pattern for prediction, we aim to involve the information of other antibiogram patterns to benefit the prediction of the target antibiogram pattern. To achieve this goal, we construct an antibiogram pattern graph $\mathcal{G}^{(k)}_t=(\mathbf{A}^{(k)}_t, \mathbf{X}^{(k)}_t)$ in $t$ for each region $r^{(k)}$. Each node in $\mathcal{G}^{(k)}_t$ represents a distinct antibiogram pattern, and each edge represents the correlation between two antibiogram patterns. 
Note that our method assumes that graph structures remain consistent for different timesteps so the antibiogram pattern graph can be simplified as $\mathcal{G}^{(k)}_t=(\mathbf{A}^{(k)}, \mathbf{X}^{(k)}_t)$. 
The antibiogram pattern graph construction consists of encoding antibiogram patterns (i.e., obtaining $\mathbf{X}^{(k)}_t$) and modeling relations between antibiogram patterns (i.e., obtaining $\mathbf{A}^{(k)}$).

\subsubsection{Encoding Antibiogram Patterns}
Before figuring out the encodings $\mathbf{X}^{(k)}_t$ for antibiogram patterns, the primary step is to convert each antibiotic $b \in \mathcal{B}$ into an encoding vector. Specifically, we use an identity matrix $\mathbf{B} \in \mathbb{R}^{M\times M}$ to denote the one-hot encodings of the antibiotics in $\mathcal{B}$ and its $m$-th row $\mathbf{b}_m$ denotes the encoding of the $m$-th antibiotic $b_m$. 

The next step is to encode antibiogram patterns. Suppose that $N$ distinct antibiogram patterns $\{\mathcal{P}_1, \mathcal{P}_2, \cdots, \mathcal{P}_N\}$ in total are ever detected in the historical data. For each antibiogram pattern $\mathcal{P}_u$, we compute its encoding $\mathbf{h}_u$ by adding the encodings $\mathbf{b}_m$ of each antibiotic $b_m$ involved by $\mathcal{P}_u$. It is worthwhile to point out that we use $-\mathbf{b}_m$ to compute $\mathbf{h}_u$ when the antibiotic $b_m$ has the resistance state $S$ in $\mathcal{P}_u$. 
Note that $\mathbf{h}_u$ is constant for the antibiogram pattern $\mathcal{P}_u$ with different timesteps since it does not include any information of $\mathcal{P}_u$'s presence in $t$ for region $r^{(k)}$. 
Considering this, we use a binary value $q^{(k)}_{t,u} \in \{0,1\}$ to represent $\mathcal{P}_u$'s presence in $t$ for region $r^{(k)}$. 
Specifically, $q^{(k)}_{t,u}=1$ when $\mathcal{P}_u$ is detected; otherwise, $q^{(k)}_{t,u}=0$. Finally, we concatenate $\mathbf{h}_u$ and $q^{(k)}_{t,u}$ to obtain $\mathcal{P}_u$'s encoding in $t$ for region $r^{(k)}$
\begin{equation}
    \mathbf{x}^{(k)}_{t,u}=[\mathbf{h}_u;q^{(k)}_{t,u}] \in \mathbb{R}^{M+1},
\end{equation}
where $[\cdot;\cdot]$ denotes the concatenation operation.

\subsubsection{Modeling Relations between Antibiogram Patterns}
In practice, antibiogram patterns may co-occur with others frequently. As a consequence, aggregating the information from other co-occurring antibiogram patterns can benefit the prediction on the presence of the target antibiogram pattern. In this work, we propose to construct an antibiogram pattern graph $\mathcal{G}^{(k)}_t$ where each node represents an antibiogram pattern and employ a GNN model which gathers information from other antibiogram patterns through links to produce an effective embedding of a target antibiogram pattern. In this scenario, the main goal is to appropriately model relations between antibiogram patterns (i.e., obtain $\mathbf{A}^{(k)}$).

A straightforward approach to achieve the goal above is to directly compute encoding similarities between antibiogram patterns as relations. However, this simple approach does not work well when similar antibiogram patterns barely co-occur. Considering the two antibiogram patterns $\{B_1, B_2, B_3, B_4\}$ and $\{B_1, B_2, B_3, B_5\}$, both of them have $B_1$, $B_2$, and $B_3$. However, the two patterns may not co-occur in the same region in practice.
Therefore, we use the Jaccard similarity \cite{yin2006jaccard} matrix $\mathbf{J}^{(k)}\in \mathbb{R}^{N\times N}$ to measure correlations between antibiogram patterns for region $r^{(k)}$ and its entry $\mathbf{J}_{uv}^{(k)} \in \mathbf{J}^{(k)}$ represents the Jaccard similarity between the antibiogram patterns $\mathcal{P}_u$ and $\mathcal{P}_v$. Let $\mathcal{S}_u^{(k)}$ denote the presence set of the antibiogram pattern $\mathcal{P}_u$ in different timesteps for region $r^{(k)}$. The Jaccard similarity $\mathbf{J}_{uv}^{(k)}$ between $\mathcal{P}_u$ and $\mathcal{P}_v$ can be computed by 
\begin{equation} \label{jaccard}
    \mathbf{J}_{uv}^{(k)}=\frac{|\mathcal{S}_u^{(k)} \cap \mathcal{S}_v^{(k)}|}{|\mathcal{S}_u^{(k)} \cup \mathcal{S}_v^{(k)}|}.
\end{equation}
Considering the example in Fig.~\ref{fig:example}, $\mathcal{S}_1^{(3)}$ of Pattern 1 is $\{1, 1, 1, 1\}$ and $\mathcal{S}_4^{(3)}$ of Pattern 4 is $\{0, 1, 1, 1\}$ for region $r^{(3)}$. Since $|\mathcal{S}_1^{(3)} \cap \mathcal{S}_4^{(3)}|=|\{1, 1, 1\}|=3$ and $|\mathcal{S}_1^{(3)} \cup \mathcal{S}_4^{(3)}|=|\{1, 1, 1, 1\}|=4$, the Jaccard similarity $\mathbf{J}_{14}^{(3)}=\mathbf{J}_{41}^{(3)}=0.75$ between $\mathcal{P}_u$ and $\mathcal{P}_v$.

For the adjacency matrix $\mathbf{A}^{(k)}$, each entry $\mathbf{A}_{uv}^{(k)}=\mathbf{J}_{uv}^{(k)}$ if $\mathbf{J}_{uv}^{(k)}>\delta$, otherwise $\mathbf{A}_{uv}^{(k)}=0$. Here $\delta$ is a hyperparameter.

\subsection{Antibiogram Pattern Graph Convolution Module}
The intuition of antibiogram pattern graph convolution is to obtain the embedding of each antibiogram pattern with respect to the encoding of the antibiogram pattern as well as those from its neighboring antibiogram patterns. Specifically, given the antibiogram pattern graph $\mathcal{G}^{(k)}_t$ in $t$ for region $r^{(k)}$, we employ a GNN model $f$ as the antibiogram pattern graph convolution module to compute the antibiogram pattern embeddings $\mathbf{Z}^{(k)}_t \in \mathbb{R}^{N\times d_1}$ by
\begin{equation}
   \mathbf{Z}^{(k)}_t = f(\mathbf{X}^{(k)}_t, \mathbf{A}^{(k)};\theta^f),
\end{equation}
where $\theta^f$ is the parameters in the GNN model $f$. In this study, we instantiate the GNN model as a two-layer GCN \cite{kipf2016gcn}.

\subsection{Temporal Attention Module}
Now we are able to make a prediction of a target antibiogram pattern $\mathcal{P}_u$ based on its embedding $\mathbf{z}^{(k)}_{t,u} \in \mathbf{Z}^{(k)}_t$ in $t$ with the antibiogram pattern graph convolution module.
%
Nevertheless, the presence of $\mathcal{P}_u$ in $T+1$ is not only dependent on its embedding $\mathbf{z}^{(k)}_{T,u}$ in $T$ but also related to its embeddings in the past several timesteps. 
Therefore, we take temporal dependencies into account and design a temporal attention module $g$ in STAPP to capture intra-region temporal dependencies. This module consists of an attention layer \cite{vaswani2017transformer} and a position-wise feed-forward network layer. Let $\mathbf{Z}^{(k)}_u \in \mathbb{R}^{T\times d_1}$ denote the stack of the embeddings $\{\mathbf{z}^{(k)}_{t,u}\}_{t=1}^T$ of $\mathcal{P}_u$ from $t=1$ to $t=T$ for region $r^{(k)}$. This module takes $\mathbf{Z}^{(k)}_u$ as the input and produces the temporal embedding $\mathbf{e}^{(k)}_u \in \mathbb{R}^{d_4}$. Specifically, it can be formulated as
\begin{equation}
   \mathbf{e}^{(k)}_u = g(\mathbf{Z}^{(k)}_u; \theta^g) = \text{FFN}(\text{Attn}(\mathbf{Z}^{(k)}_u;\theta^g_1);\theta^g_2),
\end{equation}
where $\theta^g=\{\theta^g_1, \theta^g_2\}$ denotes the parameters in $g$. $\text{Attn}(\cdot)$ represents the attention layer and $\text{FFN}(\cdot)$ represents the position-wise feed-forward network layer.

In the attention layer, we first linearly project $\mathbf{Z}^{(k)}_u$ into the query $\mathbf{Q}^{(k)}_u$, the key $\mathbf{K}^{(k)}_u$, and the value $\mathbf{V}^{(k)}_u$ through the parameters $\theta^g_1=\{\mathbf{W}_Q, \mathbf{W}_K, \mathbf{W}_V\}$ by
\begin{equation}
   \mathbf{Q}^{(k)}_u = \mathbf{Z}^{(k)}_u \mathbf{W}_Q, \mathbf{K}^{(k)}_u = \mathbf{Z}^{(k)}_u \mathbf{W}_K, \mathbf{V}^{(k)}_u = \mathbf{Z}^{(k)}_u \mathbf{W}_V,
\end{equation}
where $\mathbf{W}_Q \in \mathbb{R}^{d_1\times d'}$, $\mathbf{W}_K \in \mathbb{R}^{d_1\times d'}$, $\mathbf{W}_V\in \mathbb{R}^{d_1\times d_2}$ are learnable projection matrices and shared by all the antibiogram patterns. Then the attention layer adopts the scaled dot-product attention mechanism \cite{vaswani2017transformer} to compute attentions 
\begin{equation}
    \text{Attn}(\mathbf{Z}^{(k)}_u;\theta^g_1) = \text{softmax} \left(\frac{\mathbf{Q}^{(k)}_u \mathbf{K}_u^{(k)\top}} {\sqrt{d'}}\right)\mathbf{V}^{(k)}_u,
\end{equation}
where $\text{softmax}(\cdot)$ is the softmax operation applied in a row-wise manner. 
Typically, we instantiate a multi-head version of the attention layer by projecting $\mathbf{Z}^{(k)}_u$ into $P$ different sets of queries, keys, and values. Then we combine the $P$ attention results together and pass them into the position-wise feed-forward network layer to obtain the temporal embedding $\mathbf{e}^{(k)}_u$.

In the position-wise feed-forward network layer, we use a two-layer feed-forward network with a ReLU operation through the parameters $\theta^g_2=\{\mathbf{W}_1, \mathbf{W}_2\}$
\begin{equation}
    \text{FFN}(x;\theta^g_2) = \text{ReLU} (x\mathbf{W}_1)\mathbf{W}_2,
\end{equation}
where $\mathbf{W}_1 \in \mathbb{R}^{Td_2\times d_3}$ and $\mathbf{W}_2 \in \mathbb{R}^{d_3\times d_4}$ are learnable projection matrices.

To facilitate model training, we follow the strategy in \cite{vaswani2017transformer} and involve a residual connection \cite{he2016resnet}, layer normalization \cite{ba2016layernorm}, and positional encoding in the temporal attention module.

\subsection{Spatial Graph Convolution Module}
The significant antibiogram patterns in a region are usually related to those in other geographically close regions. We propose to utilize information from neighboring regions which are geographically close to the target region for predicting antibiogram patterns. In this study, we model the spatial dependencies based on geographical distances between regions and design an inter-region spatial graph convolution module in STAPP. Specifically, we construct a region graph $\mathcal{G}^{\mathcal{R}}$ including all the regions in $\mathcal{R}$ as nodes, and the spatial graph convolution module captures inter-region spatial correlations by applying a GNN model $h$ on $\mathcal{G}^{\mathcal{R}}$ to obtain the embeddings $\mathbf{h}^{(k)}_u$.

We first construct the adjacency matrix $\mathbf{A}^{\mathcal{R}} \in \mathbb{R}^{K \times K}$ of $\mathcal{G}^{\mathcal{R}}$ using the Gaussian kernel with a threshold \cite{meng2021cnfgnn}. Specifically, for each pair of regions $r^{(k)}$ and $r^{(q)}$, $\mathbf{A}^{\mathcal{R}}_{kq}=d_{kq}$ if $d_{kq}>\eta$, otherwise $\mathbf{A}^{\mathcal{R}}_{kq}=0$. Here $d_{kq}=\text{exp}(-\dfrac{\text{dist}(r^{(k)}, r^{(q)})^2}{\sigma^2})$ where $\text{dist}(r^{(k)}, r^{(q)})$ is the distance between region $r^{(k)}$ and region $r^{(q)}$, $\sigma$ is the standard deviation of distances, and $\eta$ is the threshold. We name our model with this construction strategy as STAPP-D.

In the meantime, we notice that geographical distances may not always determine whether antibiogram patterns are similar for two cities. Hence, we also model the spatial dependencies using the Jaccard similarity based on historical data. Specifically, for each pair of regions $r^{(k)}$ and $r^{(q)}$, we compute their Jaccard similarity $\mathbf{J}^{\mathcal{R}}_{kq}$ by Eq.~(\ref{jaccard}). $\mathbf{A}^{\mathcal{R}}_{kq}=\mathbf{J}^{\mathcal{R}}_{kq}$ if $\mathbf{J}^{\mathcal{R}}_{kq}>\kappa$, otherwise $\mathbf{A}^{\mathcal{R}}_{kq}=0$. Here $\kappa$ is a predefined threshold. We name our model with this strategy as STAPP-J.

For an antibiogram pattern $\mathcal{P}_u$, we obtain its embeddings $\{\mathbf{e}^{(k)}_u\}_{k=1}^K$ for all the regions through the temporal attention module. In the spatial graph convolution module, we employ a GNN model $h$ to compute $\mathcal{P}_u$'s final embedding $\mathbf{h}^{(k)}_u \in \mathbb{R}^{d_5}$ with respect to $\{\mathbf{e}^{(k)}_u\}_{k=1}^K$ and $\mathbf{A}^{\mathcal{R}}$
\begin{equation}
   \mathbf{h}^{(k)}_u = h(\{\mathbf{e}^{(k)}_u\}_{k=1}^K, \mathbf{A}^{\mathcal{R}};\theta^h),
\end{equation}
where $\theta^h$ is the parameters in the GNN model $h$. 

\subsection{Classifier} After obtaining $\mathcal{P}_u$'s final embedding $\mathbf{h}^{(k)}_u$ in region $r^{(k)}$, we employ a one-layer feed-forward network as the classifier $cl$ to make predictions. Specifically, the classifier module can be formulated as
\begin{equation}
   \hat{y}^{(k)}_u = cl(\mathbf{h}^{(k)}_u;\theta^{cl})=\sigma(\mathbf{h}^{(k)}_u \mathbf{w}_3),
\end{equation}
where $\theta^{cl}=\{\mathbf{w}_3\}$ is the learnable parameters of $cl$ and $\sigma$ is the sigmoid function.

\subsection{Model Training}
In this paper, we propose to formulate the problem of antibiogram pattern prediction as a node classification task. Hence, common loss functions for node classification can be adopted for training model parameters $\theta=\{\theta^f, \theta^g, \theta^h, \theta^{cl}\}$ in STAPP. In the real world, however, only a few ($\sim 2\%$) of the $N$ antibiogram patterns appear in a timestep for a region. In this scenario, node labels are significantly class-imbalanced and model parameters are prone to be biased toward major classes (i.e., ``not appear" in the problem of antibiogram pattern prediction) \cite{zhao2021graphsmote}. To tackle this, we adopt focal loss \cite{lin2017focal} as the loss function for training model parameters in STAPP. Specifically, the adopted focal loss is formulated as

\begin{equation}
\begin{aligned}
   \mathcal{L}(\theta) &= \sum_{u, k} -\lambda y^{(k)}_u (1-\hat{y}^{(k)}_u)^\gamma\log(\hat{y}^{(k)}_u)\\
   &-(1-\lambda)(1-y^{(k)}_u)
   \hat{y}^{(k)\gamma}_u \log(1-\hat{y}^{(k)}_u), 
\end{aligned}
\end{equation}
where $\lambda$ and $\gamma$ are hyperparameters.

\input{table_group.tex}

\input{table_main.tex}
\section{Experiments}
\subsection{Settings}
\subsubsection{Datasets}
We verify the effectiveness of STAPP on a real-world antibiogram dataset. This antibiogram dataset includes annual antibiogram reports from 1999 to 2012 for 203 cities in the United States, obtained from the Surveillance Network (TSN) database, a repository of susceptibility test results collected from more than 300 microbiology laboratories in the United States \cite{klein:aje13}.

These antibiogram reports include resistance states of patients for \emph{Staphylococcus aureus} to 22 distinct drugs. Note that resistance states for \emph{Staphylococcus aureus} to the drugs within a class will be consistent. Hence we group the drugs within a class based on the a priori grouping information and the number of drugs is reduced from 22 to 17. Table~\ref{table:group} shows the grouping information of the 22 drugs in our dataset. We extract $N=5,038$ distinct antibiogram patterns from the antibiogram dataset using Kingfisher \cite{hamalainen2012kingfisher} and construct antibiogram pattern graphs. We consider each year as a timestep and each city as a region in our experiments. We use the antibiogram patterns from 1999 to 2010 as the training set and those from 2011 to 2012 as the test set.

\subsubsection{Baselines}
We compare STAPP with the following six baselines. Since there are no existing studies investigating the problem in this paper, we choose several recent baselines from other domains and adapt them to this problem.

\begin{itemize}
    \item Random: This method randomly selects antibiogram patterns and predicts them as ``appear".
    
    \item LastYear: This method directly takes the presence of the antibiogram patterns in year $T$ as predictions.
    
    \item Mode: This method predicts an antibiogram pattern as ``appear" if it was detected at least $T/2$ times during the past $T$ years, otherwise we predict it as ``not appear".
    
    \item Support vector machine (SVM) \cite{lin2013svm}: It finds a boundary for classification given the historical presence of a target antibiogram pattern in the past $T$ years.
    
    \item LSTM \cite{altche2017lstm}: It is a two-layer LSTM that takes the presence of a target antibiogram pattern in the past $T$ years for a region as the input.
    
    \item T-GCN \cite{zhao2019tgcn}: It employs a two-layer GCN model for local antibiogram patterns and a two-layer GRU model for temporal dependencies.

\end{itemize}

\begin{figure*}[!t]
\centering
\subcaptionbox{$T=3$}
{\includegraphics[width=0.49\linewidth]{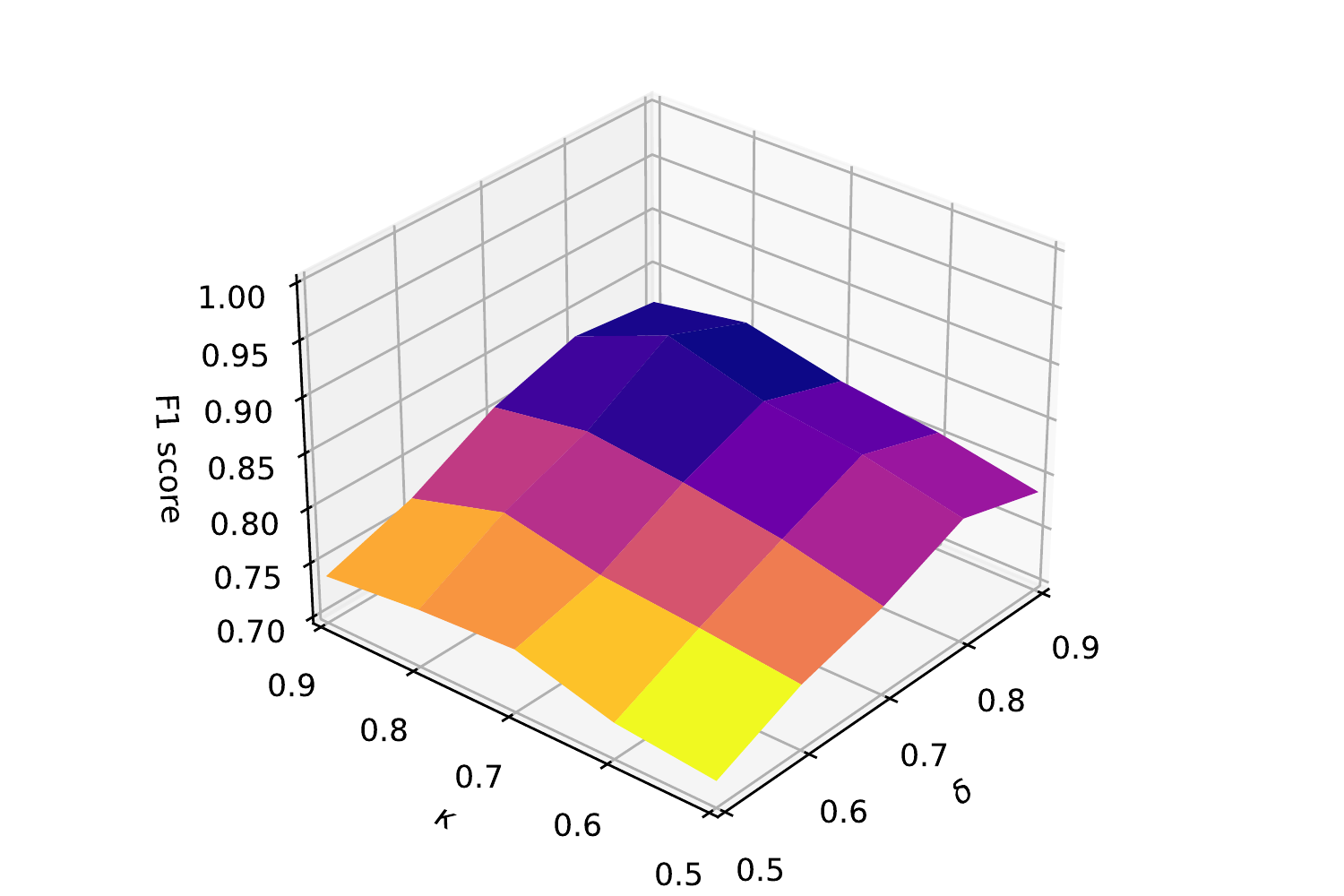}}
\subcaptionbox{$T=7$}
{\includegraphics[width=0.49\linewidth]{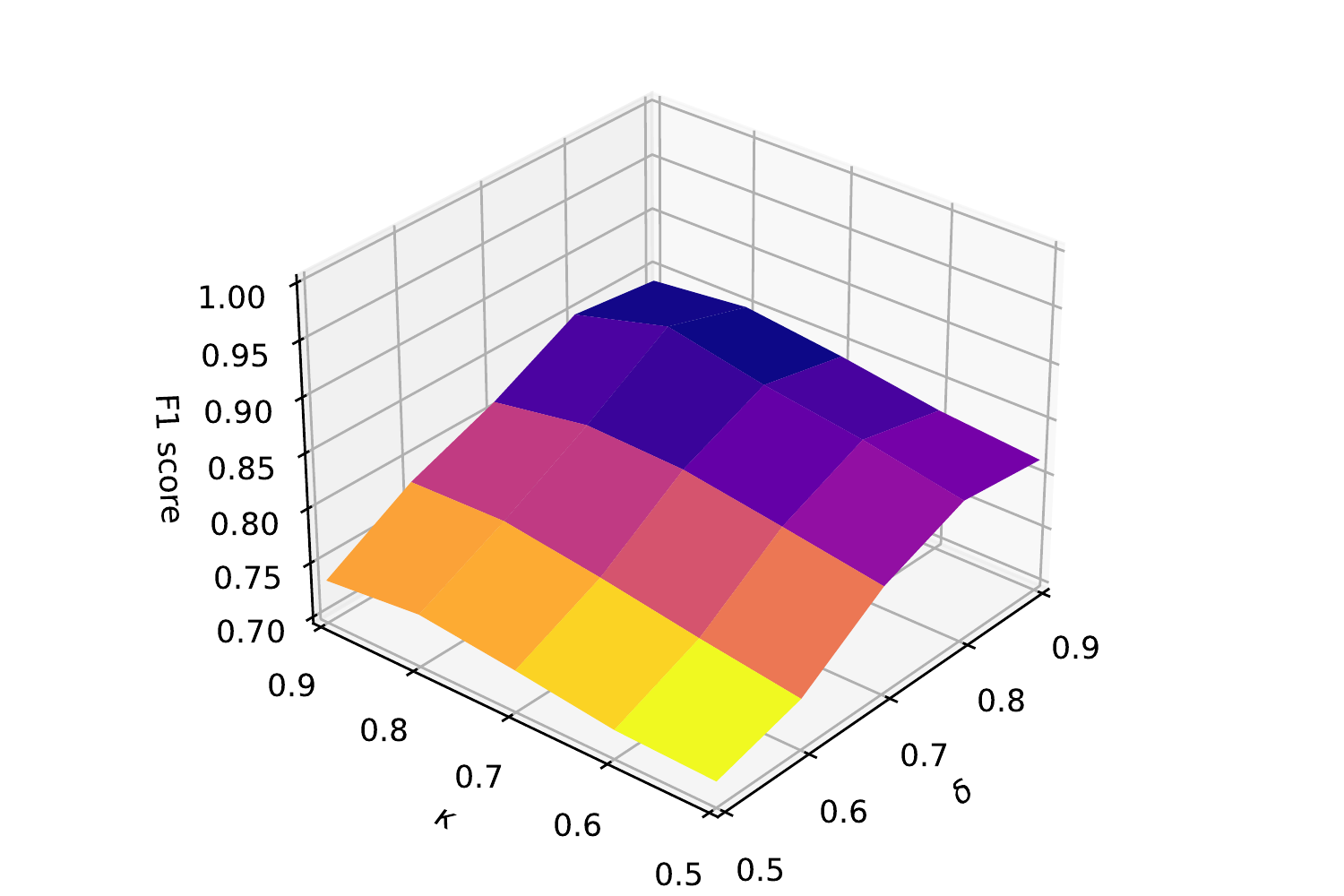}}
\caption{F1 score with different values of $\delta$ and $\kappa$.}
\label{fig:sen_f1}
\end{figure*}
\subsubsection{Experiment Setup}
As for hyperparameters during graph construction, $\delta$, $\eta$ and $\kappa$ are set as 0.8, 0.8, and 0.8, respectively. As for hyperparameters in the main modules, the hidden size of GNN models in $\theta^f$ and $\theta^h$ is set as 16. $d_1=d'=d_2=16$, $d_3=256$, $d_4=64$, $d_5=16$. We train our model using Adam optimizer with a learning rate of 0.001. The maximum training iteration is set to 500. We set $\lambda=0.7$ and $\gamma=2$ in the focal loss. $T$ is set as 3 and 7 in our experiments.

\subsubsection{Metrics} Since we formulate the problem of antibiogram pattern prediction as a node classification task, we adopt common metrics in evaluating the performance for node classification. Specifically, we use precision, recall, F1 score, AUC-ROC and NDCG@100. Note that NDCG@100 \cite{jarvelin2002ndcg} is a ranking metric that measures the similarity of ranking lists between prediction (predicted logits) and ground truth ($p$-value). To obtain the values of each metric, we first compute the results for each city and use the average values of all the cities as the final results.

\subsection{Main Results}
In this subsection, we evaluate the performance of STAPP against the other baselines and summarize the main results in Table~\ref{table:table1} and Table~\ref{table:table2} with $T=3$ and $T=7$. Here we run the experiments of each algorithm 5 times and report the average values of each metric with standard deviations. 

According to the results, we can observe the poor performance of random selection compared with other methods. Considering the imbalanced-class issue where only a small fraction of antibiogram patterns appear ($\sim2\%$) in every year, random selection will have very limited precision with around 0.02 (which is significantly smaller than 0.5 in the class-balanced setting). Therefore, the results in F1 score are extremely lower than others. On the other hand, LastYear and Mode directly utilize historical data for making predictions and can achieve comparable performance to SVM and LSTM. It suggests that antibiogram patterns may keep appearing in a city for years and involving historical data is helpful for predicting antibiogram patterns.

As for machine learning-based algorithms, SVM and LSTM only show marginal or no performance gain compared with LastYear and Mode. On the contrary, T-GCN achieves significantly higher performance than SVM and LSTM. It is because T-GCN takes antibiogram pattern graphs into account and graph structures provide abundant information on relations between antibiogram patterns. Note that we construct antibiogram pattern graphs in T-GCN using the same strategy as in STAPP. The remarkable performance gain indicates the effectiveness of our strategy. Finally, we observe that the proposed method STAPP-J achieves the best performance for all the metrics while STAPP-D has better performance than the baselines for a few metrics. For instance, STAPP-J outperforms other baselines by a large margin ($\sim 9\%$ for $T=7$) on the recall values which are much more important to evaluate the performance in the problem of antibiogram pattern prediction. In the meantime, the performance gain of STAPP-D on the recall values is about $3\%$ for $T=7$. This observation indicates that the distance-based construction strategy in STAPP-D may connect a region with improper neighboring regions and introduce noisy information to the city through their links. On the contrary, STAPP-J constructs region graphs in a data-driven way and the Jaccard similarity can appropriately reflect the similarities of antibiogram patterns between regions. 

\begin{figure*}[!t]
\centering
\includegraphics[width=0.68\linewidth]{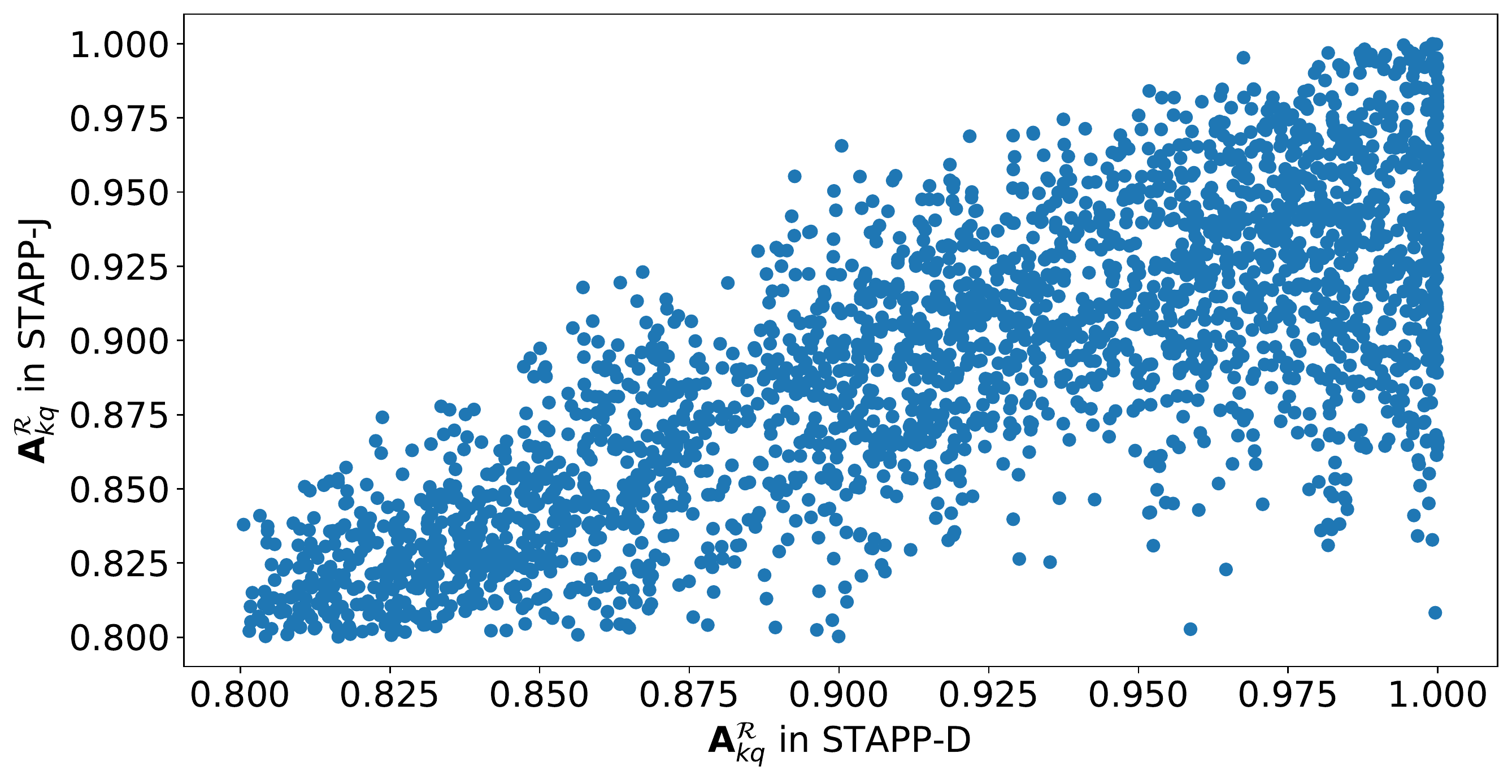}
\caption{Comparison of $\mathbf{A}^{\mathcal{R}}_{kq}$ in STAPP-D and STAPP-J.}
\label{fig:dist_sim}
\end{figure*}

\subsection{Sensitivity Study}
In this subsection, we conduct experiments to evaluate the performance of STAPP-J with different values of hyperparameters. Fig.~\ref{fig:sen_f1} shows F1 scores of STAPP-J with different values of $\delta$ and $\kappa$. We observe that $\delta$ has more effects on F1 score than $\kappa$ for both $T=3$ and $T=7$. When $\delta$ varies, F1 scores fluctuate significantly. In addition, $\delta=0.8$ and $\kappa=0.8$ are the proper choice for STAPP-J. Although larger thresholds (e.g., 0.9) can provide more similar neighbors in graphs, they may result in sparse graphs, and therefore isolated nodes are not able to obtain enough information from neighbors.

\subsection{Comparison of Distance-based and Similarity-based Regional Graph Construction Strategies}
As introduced in Section~\ref{method}, we construct the adjacency matrix $\mathcal{A}^{\mathcal{R}}$ with two strategies: the distance-based method in STAPP-D and the similarity-based method in STAPP-J. In this subsection, we aim to compare these two strategies. 
Ideally, $\mathbf{A}^{\mathcal{R}}_{kq}$ in STAPP-J will be very close to the corresponding entry in STAPP-D if $\mathcal{A}^{\mathcal{R}}$ constructed by the distance-based strategy in STAPP-D is consistent with that by the similarity-based strategy in STAPP-J. Fig.~\ref{fig:dist_sim} shows the results of $\mathbf{A}^{\mathcal{R}}_{kq}$ in STAPP-J with respect to $\mathbf{A}^{\mathcal{R}}_{kq}$ in STAPP-D. We retain each pair of $\mathbf{A}^{\mathcal{R}}_{kq}$ in STAPP-D and STAPP-J which are both larger than 0.8. From Fig.~\ref{fig:dist_sim}, we observe that the $\mathbf{A}^{\mathcal{R}}_{kq}$ in STAPP-J is generally proportional to that in STAPP-D. We compute their correlation coefficient $r=0.8129$. However, there still exist pairs of cities which are geographically close to each other (i.e, $\mathbf{A}^{\mathcal{R}}_{kq}$ in STAPP-D is close to 1) but their Jaccard similarity is not large. Therefore, the cities may not have similar prevalent antibiogram patterns which can help predict antibiogram patterns in the future. This observation indicates the necessity of the similarity-based inter-region graph construction in STAPP-J.

\section{Related Work}
\subsection{Antibiogram Patterns}
The emergence of antibiotic resistance among the most prevalent bacterial pathogens is recognized as a significant public health threat affecting humans worldwide \cite{munita2016mechanisms}. As an effective tool for detecting and monitoring trends in antibiotic resistance, antibiograms have been adopted by many infection preventionists, hospital epidemiologists, and healthcare practitioners \cite{joshi2010antibiogram,truong2021antibiogram,klein:aje13}. At the patient level, an antibiogram report is a periodic summary of antibiotic resistance testing results of a patient to selected antimicrobial drugs. Antibiograms provide comprehensive information about regional antimicrobial resistance and guide the clinician and pharmacist in selecting the best
empiric antimicrobial treatment \cite{zapantis2005nationwide}. Antibiogram patterns are significant combinations of antibiotic resistance that emerge in antibiogram reports for a region due to the spread of pathogens. Instead of investigating resistance to a single antibiotic individually, the analysis of antibiogram patterns not only can monitor antibiotic resistance trends but also can track the spread of pathogens. However, the problem in antibiogram pattern analysis (e.g., antibiogram pattern prediction) is currently unexplored, and some well-designed algorithms are urgently necessary.

\subsection{Graph-based Spatial-Temporal Prediction}
Spatial-temporal prediction plays an important role in various applications such as crowd flows prediction \cite{zhang2017crowd}, air pollution forecasting \cite{yi2018air}, and crime prediction \cite{huang2018crime}. To incorporate spatial dependencies more effectively, some recent studies investigate GNN-based approaches (e.g., GCN \cite{kipf2016gcn} and GraphSage \cite{hamilton2017graphsage}) for spatial-termporal prediction. For instance, ST-GCN \cite{yu2018stgcn} and ST-MGCN \cite{geng2019stmgcn} propose to leverage graph convolution networks to model correlations between regions. DCRNN \cite{li2018dcrnn} utilizes the bi-directional random walks on the traffic graph to model spatial information and captures temporal dynamics by GRUs. T-GCN \cite{zhao2019tgcn} leverages GCN and GRUs \cite{cho2014gru} to learn information based on topological structures and traffic dynamic patterns. Furthermore, the attention mechanism is utilized by researchers to aggregate information from adjacent roads \cite{zheng2020gman,wang2020traffic}. However, the existing studies are difficult to be directly adopted in the problem of antibiogram pattern prediction. The aforementioned methods construct graphs with respect to physical properties (e.g., road distance) while finding such relations among antibiogram patterns is very challenging in our problem. In addition, predicting the presence of an antibiogram pattern brings more difficulties, especially the class-imbalanced problem. In practice, only a small fraction of antibiogram patterns in the historical data appear in a period. Simply training a model using the cross-entropy loss may lead to predictions as ``disappear" for all antibiogram patterns. Unfortunately, to the best of our knowledge, none of the existing works are attempting to solve the above two challenges in antibiogram pattern prediction. Motivated by this, we propose STAPP which models relations between antibiogram patterns based on historical data and leverages GNN-based modules to incorporate the relations more effectively.

\section{Conclusion}
Antibiogram patterns are significant combinations of antibiotic resistance in a region. In this study, we propose a novel framework STAPP to deal with the problem of antibiogram pattern prediction. STAPP first constructs antibiogram pattern graphs by treating antibiogram patterns as nodes in a graph and modeling relations between antibiogram patterns. Then an antibiogram pattern graph convolution module is employed to aggregate information through relations between antibiogram patterns. In addition, STAPP involves a temporal attention module to capture temporal dependencies of antibiogram patterns within a region. To take the spread of antibiotic resistance into account, STAPP uses a spatial graph convolution module to extract spatial correlations among regions. We conduct extensive experiments on a real-world dataset with antibiogram reports from 203 cities in the US from 1999 to 2012. The experimental results validate the superiority of the proposed framework against other baselines.

\section*{Acknowledgments}
This work is supported by the National Center for Advancing Translational Science (UL1TR003015, KL2TR003016 to G.R.M.),  the National Science Foundation grants (IIS-2006844, IIS-2144209, IIS-2223769,
CCF-1918656, CNS-2041952, and IIS-1955797), 
the National Institutes of Health grant 2R01GM109718-07, and
the CDC MInD Healthcare network
cooperative agreement U01CK000589.
We would like to thank Saarthak Gupta (uzn2up@virginia.edu) for processing the data and running the kingfisher algorithm to find the significant patterns and Caitlyn Fay (faycaitlyn@gmail.com) for calculating the pairwise distance between cities.

\bibliographystyle{plain}
\bibliography{ref}

\end{document}

%% file: table_example.tex
\begin{table*}[!t]
\centering

\caption{An example of an antibiogram report.}
\begin{tabular}{ccccccccccc}\toprule                                                             
Amoxicillin/clavulanate     & Ampicillin        & Cefazolin         & Ceftriaxone       & Chloramphenicol           & Ciprofloxacin     & Clindamycin       & Erythromycin \\ \midrule
\emph{NULL}                 & \emph{NULL}       & \emph{S}          & \emph{S}          & \emph{NULL}               & \emph{R}          & \emph{NULL}       & \emph{R}  \\\midrule
Ampicillin/sulbactam        & Gentamicin        & Imipenem          & Levofloxacin      & Linezolid                 & Moxifloxacin      & Nitrofurantoin    & Oxacillin \\ \midrule
\emph{NULL}                 & \emph{S}          & \emph{NULL}       & \emph{S}          & \emph{NULL}               & \emph{S}          & \emph{NULL}       & \emph{S} \\ \midrule
Quinupristin/dalfopristin   & Penicillin        & Tetracycline      & Trimeth/sulfa     & Rifampin (Rifampicin)     & Vancomycin        \\ \midrule
\emph{NULL}                 & \emph{NULL}       & \emph{R}          & \emph{S}          & \emph{NULL}               & \emph{S} \\ \bottomrule
\multicolumn{3}{l}{\emph{S}: sensitive; \emph{R}: resistant; \emph{NULL}: unknown}\\
\end{tabular} 
\label{table:example}
\end{table*}

%% file: table_notations.tex
\begin{table}[!t]
\centering

\caption{Notations and the corresponding descriptions.}
\begin{tabular}{ll}\toprule
Notations                                       & Definitions and descriptions \\\midrule
$r^{(k)}$                                       & A region \\
$t$                                             & A timestep \\
$\mathcal{B}$                                   & The set of antibiotics \\
$b$                                             & An antibiotic in $\mathcal{B}$ \\
$\mathcal{P}^{(k)}_t$                           & An antibiogram pattern in $t$ for region $r^{(k)}$ \\
$\mathcal{P}_u$                                 & The $u$-th extracted antibiogram pattern \\
$\mathcal{PS}^{(k)}_t$                          & The antibiogram pattern set in $t$ for region $r^{(k)}$ \\ 
$B$                                             & A single antibiotic in $\mathcal{P}^{(k)}_t$ with its resistance state\\
$\mathcal{S}_a$                                 & The antecedent of a dependency rule \\
$\mathcal{G}^{(k)}_t$                           & The antibiogram pattern graph in $t$ for region $r^{(k)}$ \\
$\mathbf{X}^{(k)}_t$                            & The encodings of antibiogram patterns in $\mathcal{G}^{(k)}_t$ \\
$\mathbf{x}^{(k)}_{t,u}$                        & The encoding of $\mathcal{P}_u$ in $t$ for region $r^{(k)}$ \\                
$\mathbf{A}^{(k)}$                              & The adjacency matrix of all $\{\mathcal{G}^{(k)}_t\}_{t=1}^T$ for region $r^{(k)}$ \\
$\mathbf{J}^{(k)}$                              & The Jaccard similarity matrix for region $r^{(k)}$ \\
$\mathcal{S}^{(k)}_u$                           & The presence set of $\mathcal{P}_u$ for region $r^{(k)}$\\
$\mathcal{G}^{\mathcal{R}}$                     & The region graph \\
$\mathbf{A}^{\mathcal{R}}$                      & The adjacency matrix of $\mathcal{G}^{\mathcal{R}}$ \\
$f(\cdot)$                                      & The antibiogram pattern graph convolution module \\
$g(\cdot)$                                      & The temporal attention module \\
$h(\cdot)$                                      & The spatial graph convolution module \\
$cl(\cdot)$                                     & The classifier \\
$\theta^f,\theta^g,\theta^h,\theta^{cl}$        & Model parameters in $f$, $g$, $h$ and $cl$ \\
$\mathbf{z}^{(k)}_{u.t}$                        & The antibiogram pattern embedding of $\mathcal{P}_u$ \\
$\mathbf{Z}^{(k)}_u$                            & The stacked antibiogram pattern embedding of $\mathcal{P}_u$ \\
$\mathbf{e}^{(k)}_u$                            & The temporal embedding of $\mathcal{P}_u$ \\
$\mathbf{h}^{(k)}_u$                            & The final embedding of $\mathcal{P}_u$\\
\bottomrule
\end{tabular}
\label{table:notations}
\end{table}

%% file: table_group.tex


\begin{table}[!t]
\centering
\caption{Grouping information of 22 drugs in our dataset.}
\begin{tabular}{cc}\toprule
Group Name                                      & Drugs                                                                 \\ \midrule
\makecell{Beta-lactam/\\ Beta-lactamase inhibitors}           & \makecell{Amoxicillin/clavulanate, \\ Ampicillin/sulbactamr}          \\ \midrule
Penicillins                                     & \makecell{Ampicillin, \\ Penicillin}                                  \\ \midrule
Quinolones                                      & \makecell{Ciprofloxacin, \\ Levofloxacin,\\ Moxifloxacin}             \\ \midrule
Cephalosporins                                  & \makecell{Cefazolin, \\ Ceftriaxone }                                 \\ \midrule
Others                                          &  \makecell{Oxacillin, Chloramphenicol, \\ Clindamycin, Erythromycin,\\ Gentamicin, Imipenem, \\ Linezolid, Nitrofurantoin,\\  Quinupristin/dalfopristin, \\ Rifampin (Rifampicin), \\ Tetracycline, Vancomycin, \\Trimeth/sulfa}      \\ \bottomrule
\end{tabular}
\label{table:group}
\end{table}

%% file: table_main.tex
\begin{table*}[ht]
\setlength\tabcolsep{9.8pt}
\centering
\caption{Performance (Mean$\pm$Std) of STAPP and baselines ($T=3$). Bold and underlined values indicate best and second-best mean performances, respectively.}
\begin{tabular}{c ccccc} \toprule
& \multicolumn{5}{c}{\underline{\hspace{5.9cm}$T=3$\hspace{5.9cm}}}     \\ 
            & Precision                     & Recall                        & F1 score                      & AUC-ROC                       & NDCG@100   \\\midrule
Random      & $0.0184\pm0.0134$             & $0.0212\pm0.0112$             & $0.0193\pm0.0147$             & -                             & -   \\
LastYear    & $0.4149\pm0.0000$             & $0.4498\pm0.0000$             & $0.4218\pm0.0000$             & -                             & -   \\
Mode        & $0.4961\pm0.0000$             & $0.3501\pm0.0000$             & $0.3911\pm0.0000$             & -                             & -   \\\midrule
SVM         & $0.4092\pm0.0121$             & $0.4330\pm0.0184$             & $0.4105\pm0.0114$             & $0.8125\pm0.0146$             & $0.5007\pm0.0185$    \\
LSTM        & $0.4503\pm0.0158$             & $0.4399\pm0.0137$             & $0.4346\pm0.0142$             & $0.8319\pm0.0156$             & $0.5361\pm0.0096$   \\
T-GCN       & $0.7832\pm0.0104$             & \underline{$0.8171\pm0.0112$} & $0.7894\pm0.0147$             & \underline{$0.9801\pm0.0065$} & $0.7550\pm0.0073$ \\ \midrule
STAPP-D     & \underline{$0.8102\pm0.0105$} & $0.8159\pm0.0096$             & \underline{$0.8075\pm0.0124$} & $0.9774\pm0.0083$             & \underline{$0.7561\pm0.0089$} \\
STAPP-J     & $\mathbf{0.8575\pm0.0129}$    & $\mathbf{0.8794\pm0.0107}$    & $\mathbf{0.8631\pm0.0116}$    & $\mathbf{0.9903\pm0.0082}$    & $\mathbf{0.7682\pm0.0070}$ \\ \bottomrule
\end{tabular}
\label{table:table1}
\end{table*}

\begin{table*}[ht]
\setlength\tabcolsep{9.8pt}
\centering
\caption{Performance (Mean$\pm$Std) of STAPP and baselines ($T=7$). Bold and underlined values indicate best and second-best mean performances, respectively.}
\begin{tabular}{c ccccc} 
\toprule
& \multicolumn{5}{c}{\underline{\hspace{5.9cm}$T=7$\hspace{5.9cm}}}     \\ 
            & Precision                     & Recall                        & F1 score                      & AUC-ROC                       & NDCG@100  \\\midrule
Random      & $0.0185\pm0.0115$             & $0.0192\pm0.0117$             & $0.0187\pm0.0142$             & -                             & -    \\
LastYear    & $0.4149\pm0.0000$             & $0.4498\pm0.0000$             & $0.4218\pm0.0000$             & -                             & -   \\
Mode        & $0.4278\pm0.0000$             & $0.2302\pm0.0000$             & $0.2782\pm0.0000$             & -                             & -   \\\midrule
SVM         & $0.4297\pm0.0107$             & $0.3920\pm0.0135$             & $0.4019\pm0.0163$             & $0.8052\pm0.0085$             & $0.5065\pm0.0128$    \\
LSTM        & $0.4542\pm0.0141$             & $0.4128\pm0.0139$             & $0.4209\pm0.0095$             & $0.8374\pm0.0167$             & $0.5473\pm0.0104$    \\
T-GCN       & $0.7736\pm0.0120$             & {$0.8195\pm0.0089$}           & {$0.7861\pm0.0145$}           & {$0.9739\pm0.0114$}           & \underline{$0.7832\pm0.0096$} \\ \midrule
STAPP-D     & \underline{$0.8248\pm0.0136$} & \underline{$0.8482\pm0.0125$} & \underline{$0.8224\pm0.0088$} & \underline{$0.9745\pm0.0091$} & ${0.7750\pm0.0085}$ \\
STAPP-J     & $\mathbf{0.8593\pm0.0143}$    & $\mathbf{0.9114\pm0.0089}$    & $\mathbf{0.8713\pm0.0148}$    & $\mathbf{0.9920\pm0.0076}$    & $\mathbf{0.7998\pm0.0142}$ \\ \bottomrule 

\end{tabular}
\label{table:table2}
\end{table*}